\documentclass[10pt,twocolumn,letterpaper]{article}

\usepackage{iccv}
\usepackage{times}
\usepackage{epsfig}
\usepackage{graphicx}
\usepackage{amsmath}
\usepackage{amssymb}
\usepackage{booktabs}
\usepackage{algorithm}
\usepackage{algorithmic}
\usepackage{amsfonts}
\makeatletter
\@namedef{ver@everyshi.sty}{}
\makeatother
\usepackage{tikz}
\usepackage[pagebackref,breaklinks,letterpaper,colorlinks,bookmarks=false]{hyperref}
\usepackage{threeparttable}
\usepackage[capitalize]{cleveref}

\crefname{section}{Sec.}{Secs.}
\Crefname{section}{Section}{Sections}
\Crefname{table}{Table}{Tables}
\crefname{table}{Tab.}{Tabs.}

\newcommand{\figref}[1]{Fig.~\ref{#1}}

\newcommand{\figureref}[1]{Figure~\ref{#1}}
\newcommand{\tableref}[1]{Table~\ref{#1}}

\iccvfinalcopy % *** Uncomment this line for the final submission

\def\iccvPaperID{9840} % *** Enter the ICCV Paper ID here

\ificcvfinal\fi

\begin{document}

%%%%%%%%% TITLE
\title{CAFS: Class Adaptive Framework for Semi-Supervised Semantic Segmentation}

\author{Jingi Ju\thanks{equal contribution} $^{1}$ \quad Hyeoncheol Noh\footnotemark[1] $^{1}$ \quad Yooseung Wang\footnotemark[1] $^{2}$ \quad Minseok Seo$^{3}$ \quad Dong-Geol Choi\thanks{corresponding author} $^{1}$\\
$^1$Hanbat National University \quad $^2$KAIST \quad $^3$SI Analytics Inc\\
{\tt\small \{jingi.ju, hyeoncheol.noh, dgchoi\}@hanbat.ac.kr } \\
{\tt\small yswang@kaist.ac.kr \quad minseok.seo@si-analytics.ai}
}

\maketitle
%\ificcvfinal\thispagestyle{empty}\fi

%%%%%%%%% ABSTRACT
\begin{abstract}
Semi-supervised semantic segmentation learns a model for classifying pixels into specific classes using a few labeled samples and numerous unlabeled images.
The recent leading approach is consistency regularization by self-training with pseudo-labeling pixels having high confidences for unlabeled images. 
However, using only high-confidence pixels for self-training may result in losing much of the information in the unlabeled datasets due to poor confidence calibration of modern deep learning networks. 
In this paper, we propose a class-adaptive semi-supervision framework for semi-supervised semantic segmentation (CAFS) to cope with the loss of most information that occurs in existing high-confidence-based pseudo-labeling methods.
Unlike existing semi-supervised semantic segmentation frameworks, CAFS constructs a validation set on a labeled dataset, to leverage the calibration performance for each class.
On this basis, we propose a calibration aware class-wise adaptive thresholding and class-wise adaptive oversampling using the analysis results from the validation set. 
Our proposed CAFS achieves state-of-the-art performance on the full data partition of the base PASCAL VOC 2012 dataset and on the 1/4 data partition of the Cityscapes dataset with significant margins of 83.0\% and 80.4\%, respectively.
The code is available at \url{https://github.com/cjf8899/CAFS}.

\end{abstract}

\section{Introduction}

Semantic segmentation is a core task that forms the basis of many deep learning-based applications, such as autonomous driving~\cite{siam2017deep,feng2020deep,siam2018comparative,kim2023bidirectional}, aerial imagery analytics~\cite{waqas2019isaid,zheng2020foreground,xu2022feature, noh2022unsupervised, seo2023self}, and mechanical image analytics~\cite{unet,havaei2016hemis,hesamian2019deep}.
This is a per-pixel classification problem in which each pixel of the image is classified. 
In recent years, supervised semantic segmentation methods~\cite{unet, deeplabv3+, segnet, hrnet, segformer} have made remarkable progress, based on huge amounts of data~\cite{ade20k,cityscapes,coco,pascal}.
Unfortunately, supervised semantic segmentation is costly because it requires a large number of datasets for pixel-level labeling.
In most real-world scenarios, the budget for applications based on semantic segmentation  is limited. Therefore, it is not possible to label all unlabeled data. 
To solve this real-world problem, semi-supervised semantic segmentation~\cite{
c3-semiseg,pseudoseg,pc2seg,alonso2021semi,ps-mt,ael,st++,u2pl,unimatch} has been proposed.
Semi-supervised semantic segmentation aims to achieve the performance of a model trained on a fully labeled dataset using both the labeled dataset and the unlabeled dataset within the budget.

\begin{figure}[t!]
  \centering
  \includegraphics[width=1.0\linewidth]{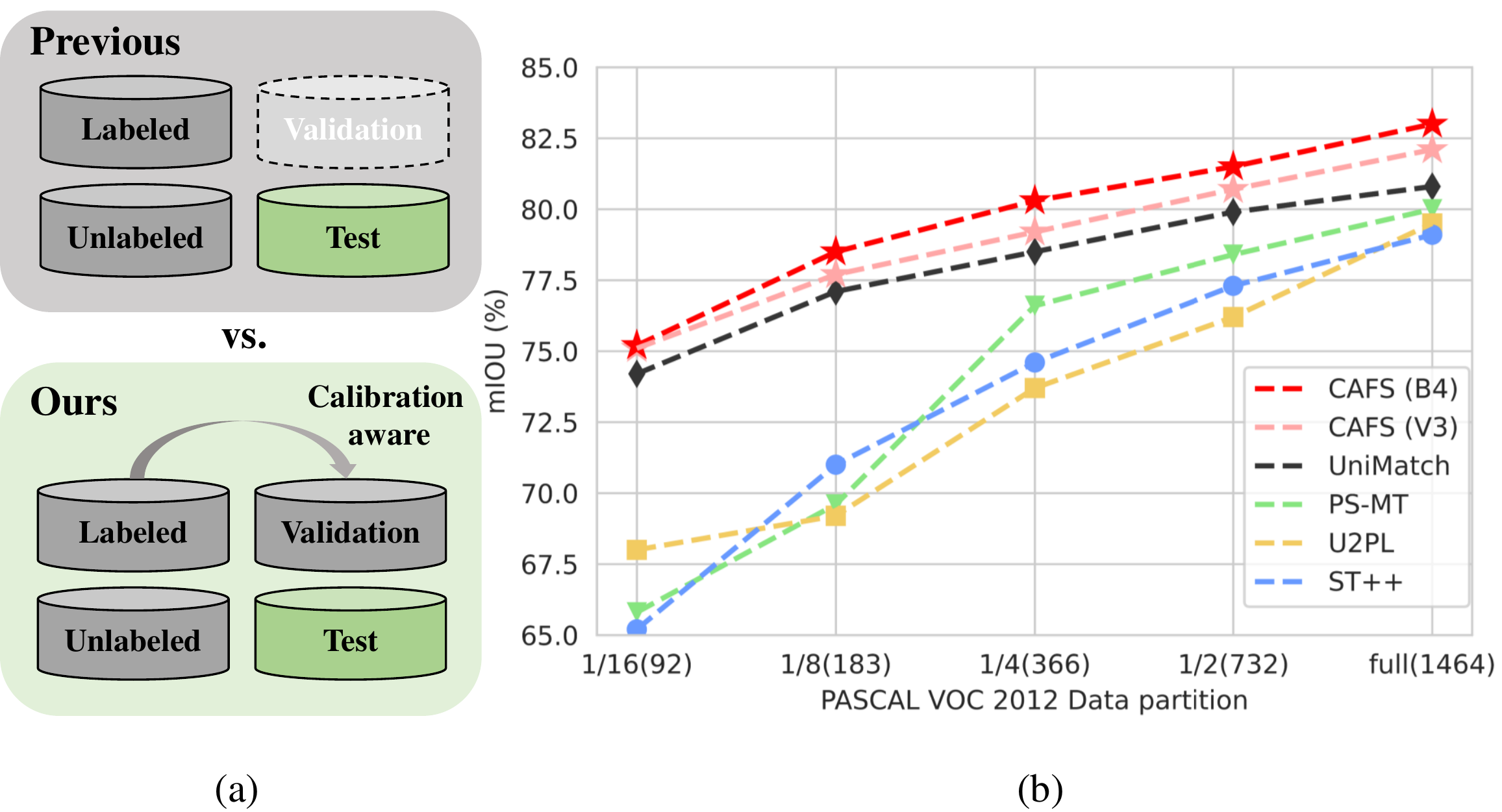}
  \caption{Performance comparison between our CAFS and its predecessor, UniMatch. (a) is a simplified diagram of the previous semi-supervised semantic segmentation framework and CAFS, and (b) is a performance comparison graph.}
  \label{fig:motivation}
  \vspace*{-0.3cm}
\end{figure}

A group of studies~\cite{u2pl, eln, em3} in semi-supervised semantic segmentation introduce entropy minimization methods to explicitly discriminate class probabilities between similar classes. 
Another group of recent studies~\cite{st++, em1, unimatch} propose methods to train robust models for various perturbations on an input image by self-training between strongly and weakly augmented images. 
They utilize a high-confidence prediction for an unlabeled data as a ground-truth label in the self-training procedure. 
Using only the high-confidence prediction has an advantage to mitigate over-fitting problem, however, they fail to address a class imbalance problem. 
Recent state-of-the-art method, UniMatch~\cite{unimatch}, used a fixed confidence threshold for training unlabeled data and cannot fully  distinguish similar classes such as \textit{motorbike} and \textit{bicycle}. 

\begin{figure}[t!]
  \centering
  \includegraphics[width=1.0\linewidth]{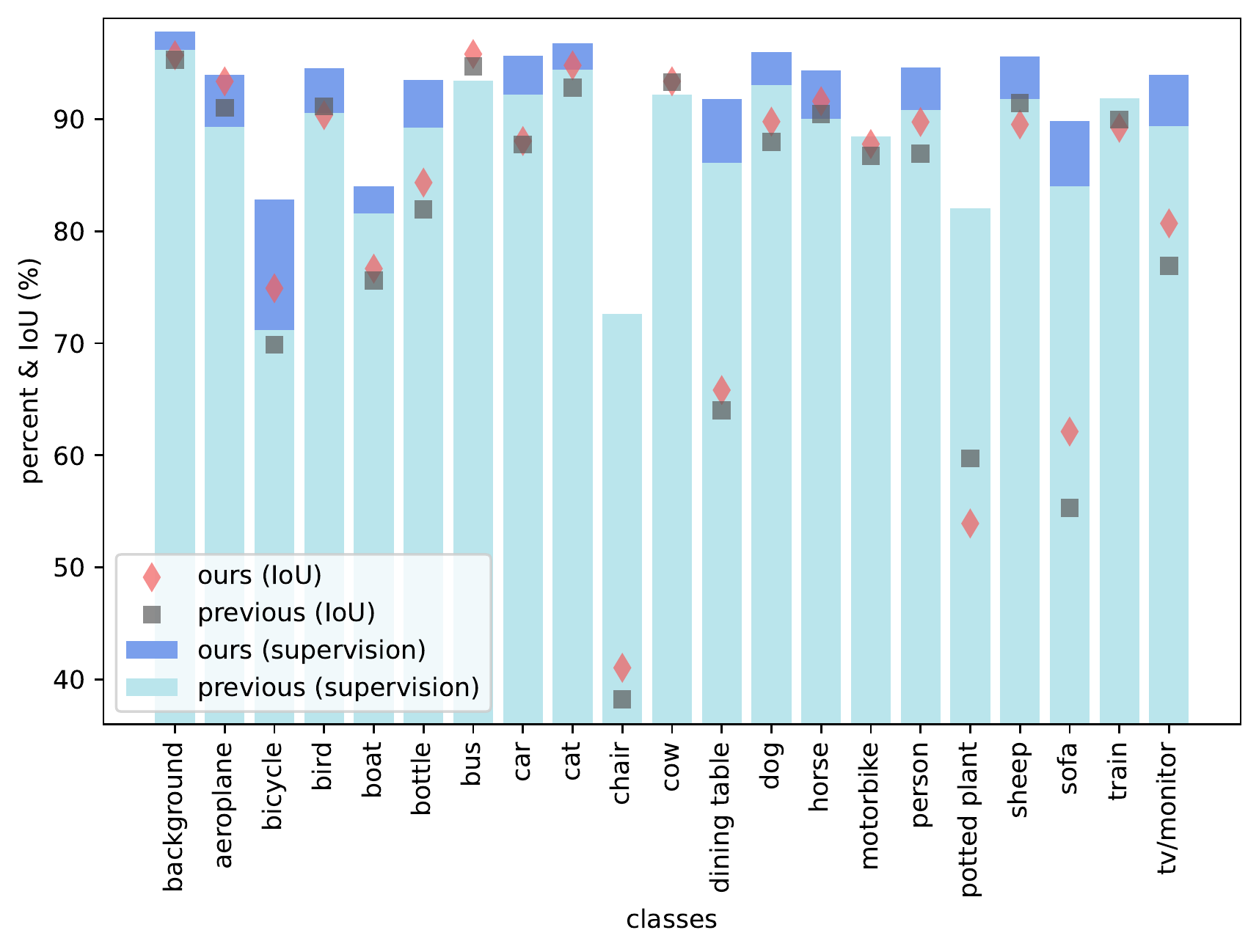}
  \caption{Class-wise performance and number of supervision pixels statistics for pseudo-label. Result of the model trained on the full (1,464) partition of the base PASCAL VOC 2012. Our (sky blue bar) and previous (mint bar) to compare the number of supervisions by class when generating pseudo-labels for unlabeled data (9,118 images). Our (diamond marker) and previous (rectangle marker) to represent the class-wise IoU of CAFS and UniMatch. By using a class adaptive threshold in the pseudo-label, we obtained better IoU results than previous work in most classes.}
  \label{fig:motivation2}
\end{figure}

In this paper, to address the class imbalance problem, we propose a class-adaptive framework for semi-supervised semantic segmentation (\textbf{CAFS}). 
In supervised semantic segmentation, the validation set is configured and its performance is exploited.
However, in the context of semi-supervised semantic segmentation, the validation set is not configured because the number of labeled datasets is small.
CAFS is a new semi-supervised semantic segmentation framework that constructs a validation set and uses the analysis results regardless of the number of labeled data.
CAFS consists of adaptive class-wise oversampling (AOS) and adaptive class-wise confidence thresholds (ACT) based on the prediction results of the validation set.
ACT can directly measure the calibration performance of the model trained on the labeled dataset, and based on that, analyze the thresholds with the optimal precision for each class and apply the optimal pseudo-label threshold on the unlabeled dataset.
~\figref{fig:motivation2} shows a graph of the ratio of pseudo-label supervision with and without ACT on the UniMatch baseline.
As shown in ~\figref{fig:motivation2}, regardless of the calibration performance of the network, UniMatch uses only high-confidence prediction as a pseudo-label, so the ratio of supervision for each class varies greatly.
However, when our ACT is applied, it can be seen that the ratio of supervision increases in most classes.

To evaluate CAFS, we used the DeepLabV3+~\cite{deeplabv3+} architecture, which has been previously used in semi-supervised semantic segmentation, and the SegFormer~\cite{segformer}, a transformer-based semantic segmentation architecture that has recently attracted much attention.
The PASCAL VOC 2012~\cite{pascal} and Cityscapes~\cite{cityscapes} datasets were used as the evaluation datasets.
As shown in ~\figref{fig:motivation}, CAFS achieved state-of-the-art performance with a large gap in various architectures and benchmark datasets compared to the previous state-of-the-art methods despite the simple modification of the existing semi-supervised semantic segmentation pipeline.

In summary, our main contributions can be described as follows:
\begin{itemize} 
\item We propose CAFS, a new framework that constructs and uses a validation set to measure the calibration performance of the network.

\item We propose ACT, which mitigates the ineffective pseudo-labeling method that uses only high-confidence prediction for all classes equally, and a performance-based AOS that does not rely on data statistics.

\item Our CAFS achieved top performance by a wide margin in all benchmarks on PASCAL VOC 2012 and Cityscapes, not only in CNN-based DeepLabV3+ but also in SegFormer, a transformer family.

\label{sec:intro}
\end{itemize} 

\section{Related Work}

\begin{figure*}[!h] 
  \includegraphics[width=\linewidth]{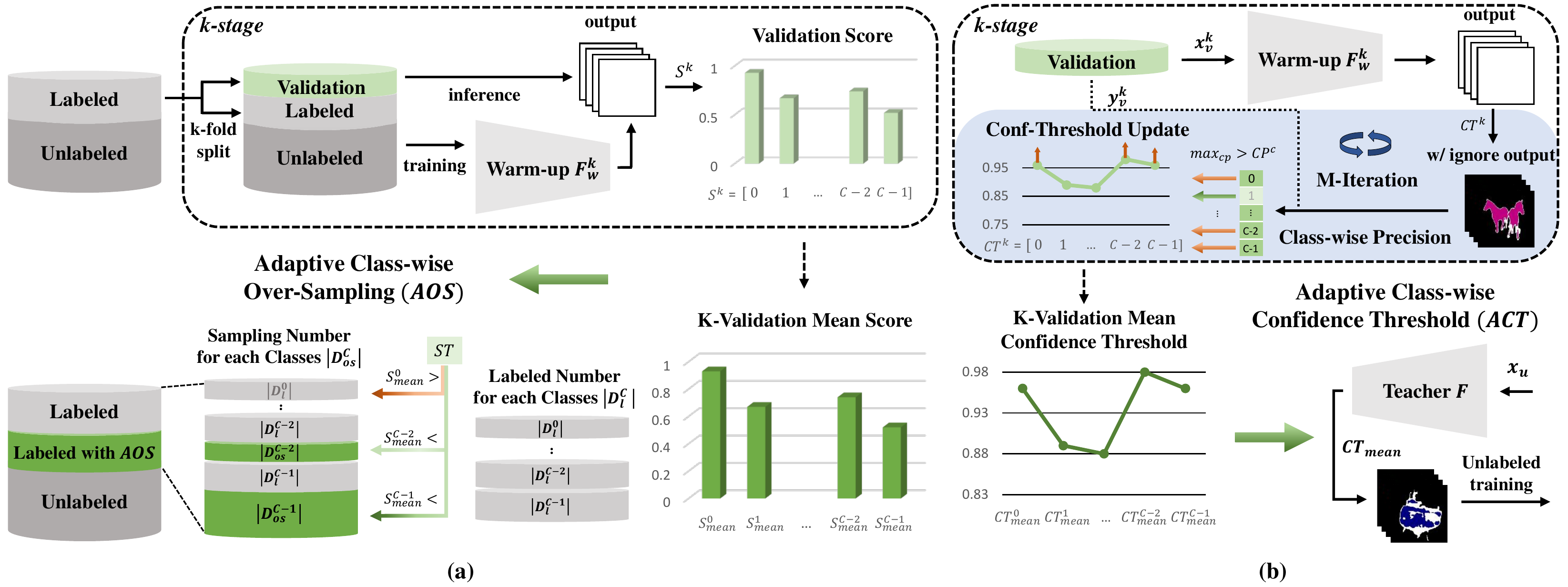}
  \caption{\textbf{Overview of the proposed AOS, ACT}. (a) Configure K number of validation sets and train a warm-up model $F_{w}^{k}$ on $D_{l-v}^{k}$ and $D_{u}^{k}$. Next, extract the K number of validation scores $S^{k}$, and calculate $S_{mean}$ by averaging each class. We perform adaptive class-wise oversampling (AOS) using the $S_{mean}$ value for each class and the sampling threshold ($ST$) as the mean of all class values in $S_{mean}$. (b) Iterate M times, and update the precision-based confidence threshold with the validation set $D_{v}^{k}$ using the trained warm-up model $F_{w}^{k}$. Then, extract K confidence thresholds $CT^{k}$ and calculate $CT_{mean}$ by averaging each class. $CT_{mean}$ is used as an adaptive class-wise confidence threshold (ACT) when generating pseudo-labels for unlabeled data.}
\label{fig:overview_images}
\end{figure*}
\label{sec:rel}
\subsection{Supervised Semantic Segmentation}
Semantic segmentation is a pixel-level classification method. 
In particular, FCN~\cite{fcn} is a fundamental task in semantic segmentation and solves semantic segmentation by pixel-to-pixel classification in an end-to-end manner.

Since then, the performance of supervised semantic segmentation has greatly improved by using transfomer-based architecture~\cite{segformer,setr}, loss function, and large-scale pre-trained weights, such as ImageNet-21k~\cite{imagenet}. 
However, in the field of semi-supervised semantic segmentation, advances in supervised semantic segmentation has not yet been exploited.
In this paper, we perform additional experiments with a transformer-based SegFormer~\cite{segformer}.

\subsection{Semi-Supervised Semantic Segmentation}
Semantic segmentation, which requires pixel-level annotation, is costly.
In real-world scenarios, unlabeled datasets can be easily used, and using them for training with unlabeled datasets helps improve the performance of the semantic segmentation model.
A recent approach that leads to semi-supervised semantic segmentation is the use of pseudo-labels~\cite{dars, ael, st++,u2pl,unimatch} of unlabeled datasets.
ST++~\cite{st++} augmented the datasets to reduce overfitting and explicitly separated them using a self-training method.
However, ST++ inevitably contains unreliable pixel values, leading to fatal performance degradation.
To address this problem, $\text{U}^2\text{PL}$~\cite{u2pl} and UniMatch~\cite{unimatch} used entropy-based and confidence-based thresholds to discriminate between reliable and unreliable pixels in pseudo-labels.
However, $\text{U}^2\text{PL}$ and UniMatch do not consider the class imbalance problem in pseudo-labels.
DARS~\cite{dars} uses the frequency counts of the real labels and pseudo-labels of each class to calculate the threshold for each class, thus reducing the number of meaningful pixels.
AEL~\cite{ael} uses the confidence bank to augment the data with adaptive copy-paste and cutmix for under-performing classes, and uses dynamic re-weighting to reduce noise from pseudo-labels. 
In this paper, we propose the following: a class adaptive threshold to efficiently generate pseudo-labels in the presence of class imbalance situations.
\subsection{Calibration of Modern Neural Networks}
Deep learning networks have poor calibrations~\cite{guo2017calibration} performance.
In particular, the calibration performance as a function of the class balance and class difficulty of the training dataset has deteriorated.
Therefore, applying the same threshold to all classes in self-training methods can potentially exacerbate the class-imbalance problem.
However, it is essentially impossible to find the optimal threshold based only on sample volume~\cite{daformer} or data statistics~\cite{dars}.
In this paper, we created a validation set from a labeled dataset and mitigated the class imbalance problem with a class-adaptive threshold using the precision of the validation set.

\section{Methods}

\label{sec:met}

\subsection{Problem Definition} 
First, we introduce the basic semi-supervised semantic segmentation settings and methods.
To train the segmentation model $F$, we have a small number of labeled datasets, $\mathcal{D}_{l} = \{ (x_{l}^{i},y_{l}^{i})\}$, and numerous unlabeled datasets, $\mathcal{D}_{u} = \{ (x_{u}^{i})\} $, where $D_l\in\mathbb{R}^{N_l}$ , $D_u\in\mathbb{R}^{N_u}$.
The goal of semi-supervised semantic segmentation is to achieve fully supervised semantic segmentation performance.
Self-training has recently become a leading approach.
Self-training generates pseudo-labels $\hat{y}_{u}$ in the unlabeled dataset $\mathcal{D}_{u}$ with the model $F$ trained on a small number of labeled datasets.
After that, the pseudo-labeled dataset $\mathcal{D}_{u}$ and the labeled dataset $\mathcal{D}_{l}$ are trained simultaneously.
CAFS first measures the calibration performance and accuracy of the current model $F$ in $\mathcal{D}_{v}$, which is a subset of the training set, and then creates an adaptive class-wise oversampled dataset $\mathcal{D}_{l}$ based on the $\mathcal{D}_{v}$ performance, and uses adaptive class-wise confidence thresholds (ACT) to generate pseudo-labels for $\mathcal{D}_{u}$.
Then, $\mathcal{D}_{l}$ and $\mathcal{D}_{u}$ were trained together.

\subsection{Motivation of proposed CAFS}
A potential problem with the existing self-training framework is that it does not take into account the calibration performance of each class.
For example, if only predictions with confidence higher than 0.95 are used for self-training, as is the case with the existing frameworks, the supervision of rare classes or difficult classes will be reduced.
Moreover, for a class like \textit{cat}, 0.98 is the threshold at which the highest accuracy can be achieved. However, at 0.95, there is a risk of overfitting the label noise and degrading the performance. 
The most intuitive and easiest way to solve this problem is to measure the calibration performance of each class.

We found that there is a strong positive correlation between the calibration performance in the randomly extracted subset from $\mathcal{D}_{l}$ and the test set.
Based on these findings, we constructed $\mathcal{D}_{v}$, which is a subset of $\mathcal{D}_{l}$, to obtain the optimal pseudo-label threshold and measure the calibration performance of each class.
Detailed experimental results can be found in the supplementary material.

\subsection{Class Adaptive Scores from Credible Sources} 
The major advantage of AOS and ACT is that they generate two scores, a class-wise balanced dataset and class-adaptive confidence threshold, from credible samples with annotations.
We create these in three stages. First, we created a validation dataset and used it to train a warm-up model.
Finally, we extracted class-adaptive confidence values and scores, as well as balanced datasets from the warm-up model and validation datasets.

Given the limited labeled dataset, $\mathcal{D}_{l}$, previous studies~\cite{unimatch, u2pl} trained the final model using an entire labeled dataset.
Splitting the validation dataset from $\mathcal{D}_{l}$ was challenging due to the limited number of samples in this dataset.
Splitting a few samples degrades the model performance. However, we solved this problem by training a warm-up model with the K-fold validation dataset.
The naive validation split generates a two-fold validation dataset $\mathcal{D}_{v}$ from the labeled dataset $\mathcal{D}_{l}$. We denote the labeled dataset as $\mathcal{D}_{l} = {(x^i, y^i)} \in \mathbb{R}^{N_{l}}$, where $N$ denotes the number of samples. 
$x^i$ and $y^i$ denote the input and label of the sample, respectively. Consider the validation dataset, $\mathcal{D}_{v}$, and a residual dataset, $\mathcal{D}_{l-v}$ as a subset of the $\mathcal{D}_{l}$, $\mathcal{D}_{v}$, $\mathcal{D}_{l-v} \subset \mathcal{D}_{l}$, then the validation dataset is formulated as follows:
\begin{equation}
  \mathcal{D}_{v} = \left\{ (x_{v}^{i},y_{v}^{i}) \right\}, \mathcal{D}_{l-v} = \left\{ (x_{l-v}^{i},y_{l-v}^{i})\right\},
  \label{eq:labeled_val_dataset}
\end{equation}
where $\mathcal{D}_{v} \in \mathbb{R}^{N_v}$, and $\mathcal{D}_{l-v} \in \mathbb{R}^{N_{l-v}}$. $x_v^i$ and $y_v^i$ are the image and label samples in the validation dataset, respectively. 
$N_v$ represents the sample numbers in the validation split and $N_{l-v} = N_{l} - N_{v}$.
Note that we extracted a relatively small portion of the validation dataset from the entire labeled dataset, $|\mathcal{D}_{v}| << |\mathcal{D}_{l}|$. 
We randomly selected the validation samples; however, we set a constraint to contain at least one image sample corresponding to all classes to extract class-aware scores from the $\mathcal{D}_{v}$.

Nevertheless, training the warm-up model with the naive validation dataset inevitably fails to predict the real-world image distribution because of the limited number of labeled datasets to understand the complex segmentation context information.
To overcome this, we propose the construction of K-fold validation datasets to improve the generalization ability using few-labeled images in the validation dataset $\mathcal{D}_{v^k} \in \mathbb{R}^{N_v}$, as follows:
\begin{equation}
  \mathcal{D}_{v}^{k} = \left\{ (x_{v}^{(k,i)},y_{v}^{(k,i)})\right\} , \mathcal{D}_{l-v}^{k} = \left\{ (x_{l-v}^{(k,i)},y_{l-v}^{(k,i)})\right\},
  \label{eq:labeled_val_dataset_v2}
\end{equation}
where $x_{v}^{(k,i)}$ and $y_{v}^{(k,i)}$ represent an image and label in the k-fold validation dataset, respectively.
Compared to Eq.~\ref{eq:labeled_val_dataset}, we keep the number of individual validation datasets at $N_{v}$, where $|\mathcal{D}_{v^0}| = |\mathcal{D}_{v^1}| = ... = |\mathcal{D}_{v}^{K-1}| = N_{v}$.
Note that the samples in each validation dataset are uniquely sampled without replacement, and thus, $\mathcal{D}_{v}^{i} \cap \mathcal{D}_{v}^{j} = \emptyset$ for all $1 \leq i \leq N_v$, $1 \leq j \leq N_v$ and $i \neq j$.

For each validation set, we trained the k-th warm-up model $F_w^{k}$.
Using the warm-up models and the K-th validation datasets, we extracted the intersection-over-union scores, $S^k$ for an AOS operation:
\begin{equation}
  S^{k} = \phi(F_{w}^{k}(x_{v}^{k}),y_{v}^{k}),
  \label{eq:basic loss}
\end{equation}
where $\phi(\cdot,\cdot)$ is an IoU function dividing overlap area by the union area. 
$S^k = \{S^k_i|i=1,2,...,C\}$ contains $C$ values, where each value represents the IoU score for the corresponding class.
Each $S^k_i$ is in the range of $0\leq S^k_i \leq  1$. To improve generalization ability, we calculated the mean value of each class, $S_{mean}$, as follows:
\begin{equation}
  S_{mean} = \frac{1}{K}\sum_{k=1}^{K} S^{k} ,
  \label{eq:basic loss}
\end{equation}
we also get the total mean over all classes, $ST$,
\begin{equation} 
ST =\frac{1}{C} \sum_{c=0}^{C-1}S^{c}_{mean} ,
\end{equation}

\subsection{Adaptive class-wise Over-Sampling}

Previous works~\cite{st++, unimatch} have attempted to improve pseudo-label quality by strong data augmentation (SDA).
However, they failed to adjust the parameters for class-wise augmentation.
UniMatch~\cite{unimatch} contains the number of images in each class.
Unlike previous studies, we propose an IoU-based adaptive class-wise oversampling method that uses a more credible validation dataset, called adaptive class-wise oversampling (AOS). 
The AOS selects hard classes based on the validation IoU scores, as follows:
\begin{equation}
    \left|\mathcal{D}_{os}^{c}\right| =\left\{\begin{matrix}  \, \left|\mathcal{D}_{l}^{c}\right| \left \lceil \lambda\,(ST-S^{c}_{mean}) \right \rceil, & \mathrm{if }\;(ST> S^{c}_{mean}),\\
     \\0, & \mathrm{otherwise},
    \end{matrix}\right.
  \label{eq:os dataset}
\end{equation}
where $|\mathcal{D}_{os}^c|$ and $|\mathcal{D}_l^c|$ are oversampled and labeled datasets for class $c$. $\lambda$ is a hyper-parameter. 
Note that the oversampling dataset, $\mathcal{D}_{os}$, is generated by using the mean IoU scores in the validation set.
The class image datasets, $D_{os}^c$  with low validation IoU score is oversampled.
Hence, the number of images with low and high IoU score images are balanced by the parameter $\lambda$.
Conducting the class-wise over-sampled dataset, $\mathcal{D}_{os}$, enables the train model to escape the class imbalance problem. 

\subsection{Adaptive class-wise Confidence Thresholds}
Compared to previous approaches that a fixed threshold method per class, we propose precision-based adaptive class-wise confidence thresholds extracted from the credible validation dataset.

Given an input image xiv in the validation set $\mathcal{D}_{v}$, we consider a pixel-wise class probability $p^i \in \mathbb{R}^{C \times H \times W}$ calculated with a softmax function.
The pseudo-label $\hat{y}^{i}_v \in \mathbb{R}^{H\times W}$ of the image is obtained by distinguishing the ignore pixel from the credible pixel in $p^i$ as follows:

\begin{equation}
\centering
    \hat{y}^{i,j}_v = 
    \begin{cases}
        {\arg\underset{c}\max} (p^{i,j}), & \textrm{if} \quad \max(p^{i,j}) > CT^c,\\
        \mathrm{ignore}, & \mathrm{otherwise,}
    \end{cases}
    % ,  c \in \{0....C-1\}
    \label{eq:Pseudo label}
\end{equation}
where $p^{i,j}\in \mathbb{R}^C$ indicates the confidence of the $j$ th pixel of image $i$.
The confusion matrix between $\hat{y}^{i}_v$ and the label of image $y^i_v$ yields a class-wise true positive (TP) and a class-wise false positive (FP). 
We update the confidence threshold (CT) per class, $CT^c$, with a class-wise precision (CP) calculated by TP and FP for all validation images. 
The update operation per class is described as follows:

\begin{equation}
\centering
    CT^{c} = 
    \begin{cases}
        CT^c + \epsilon & \mathrm{if} \quad CP^c  < max_{cp},\\
        CT^c & \mathrm{otherwise},
    \end{cases}
    % ,  c \in \{0....C-1\}
    \label{eq:threhold update}
\end{equation}
where $c \in \{0,1,… ,C-1\}$ represents the class index, and $max_{cp}$ is the precision threshold. $\epsilon$ is set to 0.01.
The $CT$ update is run $M$ times.
$M$ was set to $({max_{ct}} - {min_{ct}}) * 100$, where $max_{ct}$ and $min_{ct}$ represent the maximum and minimum values of $CT$, respectively, which were set differently depending on the number of labeled images.
The pseudo-code of ACT is described in Algorithm~\ref{alg:framework}. Thus, we have confidence thresholds for all classes, $CT^c$ to produce high-quality pseudo-labels.

\section{Experiments}

\begin{table*}[t]
\centering
\caption{Comparison with state-of-the-art for the base PASCAL VOC 2012 dataset. Previous methods and ours (V3) use a DeeplabV3+ network with ResNet101 backbone, and ours (B4) uses a SegFormer network with MiT-B4 backbone. SupOnly is a model trained with supervised without using unlabeled data. The best results are in \textbf{bold}, and the second best are \underline{underlined}. † : This means that we reproduced the approach.}
\vspace{0.1cm}
\resizebox{0.68\textwidth}{!}{
\normalsize
\begin{threeparttable}
\def\arraystretch{0.9}
\begin{tabular}{c|ccccc}
\toprule
Method        & 1/16 (92)  & 1/8 (183) & 1/4 (366) & 1/2 (732) & Full (1464) \\
\midrule
SupOnly     & 45.1       & 55.3       & 64.8       & 69.7       & 73.5       \\
\midrule
PseudoSeg~\cite{pseudoseg}     & 57.6      & 65.5       & 69.1       & 72.4       & 73.2       \\
PC$^{2}$\text{Seg~\cite{pc2seg}}        & 57.0       & 66.3       & 69.8       & 73.1       & 74.2       \\
CPS~\cite{cps}           & 64.1       & 67.4       & 71.7       & 75.9       & -          \\
ST++~\cite{st++}          & 65.2       & 71.0       & 74.6       & 77.3       & 79.1       \\
U$^{2}$\text{PL~\cite{u2pl}}           & 68.0       & 69.2       & 73.7       & 76.2       & 79.5       \\
PS-MT~\cite{ps-mt}         & 65.8      & 69.6      & 76.6      & 78.4      & 80.0      \\
% UniMatch~\cite{unimatch}      & \underline{75.2}       & 77.2       & 78.8       & 79.9       & 81.2       \\
UniMatch$\dagger$~\cite{unimatch}      & 74.2       & 77.1       & 78.5       & 79.9       & 80.8       \\
\midrule
Ours(V3)      & \underline{75.1}    & \underline{77.7}  & \underline{79.2}    & \underline{80.7}    & \underline{82.1} \\
Ours(B4)   & \textbf{75.2}   & \textbf{78.5}   & \textbf{80.3}    & \textbf{81.5}   & \textbf{83.0} \\
\bottomrule%
\end{tabular}%
\end{threeparttable}%
}%
\label{tb:ex1}%
\end{table*}%

\subsection{Setup}

\begin{algorithm}[h]
\caption{Adaptive class-wise Confidence Thresholds}\label{algorithm}
$CT$: class-wise confidence threshold;
$C$: number of classes; 
$min_{ct}$: confidence threshold minimum;
$M$: number of threshold updates; 
$TP$: class-wise true positive; 
$FP$: class-wise false positive;
$CP$: class-wise precision; 
$\mathcal \{(x^{i}_v,y^{i}_v)\}$: validation set $\mathcal{D}_{v}$;
$F_{w}$: warm-up model; 
$p^{i}$: pixel-wise class probability; 
$\hat{y}^i_v$: pseudo-label; 
$N_v$: number of validation images; 
$TP^i$: class-wise true positive per image; 
$FP^i$: class-wise false positive per image\\
\textbf{Procedure}:
\begin{algorithmic}[1]
\STATE
$CT_0=$ $...=$ $CT_{C-1}=$ $min_{ct}$
\FOR{$(m = 1; m \le M; m = m  +1)$}
    \STATE
    $TP_0=$ $...=$ $TP_{C-1}=$ 0
    \STATE
    $FP_0=$ $...=$ $FP_{C-1}=$ 0
    \STATE
    $CP_0=$ $...=$ $CP_{C-1}=$ 0
    \FOR{$(i = 1; i \le N_{v}; i = i  +1)$}
      \STATE
      $p^{i}$ = Softmax$(F_w(x^i_v))$
      \STATE
      $\hat{y}^i_v$ = Pesudo label$(p^i, CT)$
      \STATE
      $TP^i, FP^i$ = Confusion Matrix$(\hat{y}^i_u,y^i_v)$
      \STATE
      $TP$ = $TP + TP^i$
      \STATE
      $FP$ = $FP + FP^i$
    \ENDFOR{}
    \STATE
    $CP = TP / (TP+FP)$
    \STATE
    $CT$ = Update$(CP,CT)$
\ENDFOR{}
\end{algorithmic}
\label{alg:framework}
\end{algorithm}

\paragraph{Backbone network}
DeepLabV3+~\cite{deeplabv3+} backbone network is a widely used backbone network in SSL.
Transformer-based methods have recently been proposed for domain adaptation~\cite{segformer,daformer,hrda}.
We have performed experiments with both DeepLabV3+ and the transformer-based SegFormer model.
CAFS with both backbone networks achieved state-of-the-art performance.
Following previous studies, we used ResNet50 and ResNet101~\cite{resnet} for DeepLabV3+.
For transformers, we performed experiments with a Mix Transformer B4 (MiT-B4)~\cite{segformer} on SegFormer.

\paragraph{Dataset}
The PASCAL VOC 2012 dataset is a standard semantic segmentation benchmark with 20 different classes and a total of 21 classes, including a background class.
Two versions of the dataset were used.
The baseline PASCAL VOC 2012 dataset was the original dataset with 1,464 labeled images, and the extended PASCAL VOC 2012 dataset was an extended dataset with 10,582 labeled images created with lower-quality annotations using the SBD~\cite{sbd} dataset.
We evaluated both datasets and used split files that were pre-split from $\text{U}^2\text{PL}$ and UniMatch. 
Cityscapes is a dataset for understanding city scenes, with 2,975 fine annotations and images and 500 test images.

\paragraph{Training Objectives}  
In our experimental settings, we used cross-entropy ($CE$) loss.
We trained models with a hyper-parameter $\alpha$, which is a trade-off between the supervised loss $\mathcal{L}_{l}$ and unsupervised loss $\mathcal{L}_{u}$, represented as $\mathcal{L} = \mathcal{L}_{l}+\alpha \mathcal{L}_{u}$.

\paragraph{Implementation Details}  
In the DeeplabV3+ network, basic settings such as optimizer, mini-batch, hyper-parameters, input resolution, and data augmentation use the UniMatch settings as a baseline.
In the SegFormer network, if the input resolution is 312, the resolution is maintained by adjusting the stride of the last embedding block of the MiT from 2 to 1.
In the Cityscapes dataset, the optimizer used is from AdamW~\cite{adamw}, a learning rate of 6 $\times$ $10^{-5}$, and both the other settings and PASCAL VOC 2012 are the same as DeeplabV3+.
In addition, the precision threshold $max_{cp}$ was set to 0.95 in all experiments.
$\lambda$ was set to 1, and $N_v$ was set to approximately 20\% of the labeled dataset.

\subsection{Comparison with State-of-the-art Methods}

We extensively compared our method with the recent semi-supervised semantic segmentation methods: PseudoSeg~\cite{pseudoseg}, $\text{PC}^{2}\text{Seg}$~\cite{pc2seg} CPS~\cite{cps}, AEL~\cite{ael}, CAC~\cite{cac}, ST++~\cite{st++}, $\text{U}^{2}\text{PL}$~\cite{u2pl}, UniMatch~\cite{unimatch}.
The existing methods use DeepLabV3+, and only our method introduces SegFormer~\cite{segformer}.

\paragraph{PASCAL VOC 2012 Results}
~\tableref{tb:ex1} shows the comparison results for the base PASCAL VOC 2012 dataset.
CAFS outperformed all previously proposed semi-supervised semantic segmentation methods. 
Moreover, CAFS trained on the significantly low labeled partition (1/16) performed even better than the baseline, PseudoSeg~\cite{pseudoseg}, and PC$^2$Seg~\cite{pc2seg} using 16 times larger labeled samples.
Our proposed CAFS shows better performance even in the 1/8 partition, achieving 4.2\% higher mIoU than the SupOnly baseline setting trained on 1,464 samples (full). 
In all data partitions, CAFS outperformed the previous state-of-the-art method UniMatch by +0.9\%,+0.6\%,+0.7\%,+0.8\% and +1.3\% under 1/16, 1/8, 1/4, 1/2, and full partition protocols, respectively.

\begin{table}[!t]
\centering
\caption{Comparison with the state-of-the-art for the Cityscapes dataset. Existing methods and ours (V3) use the DeeplabV3+ network and ResNet101 backbone, and ours (B4) uses SegFormer network and the MiT-B4 backbone.}
\vspace{0.1cm}
\resizebox{0.4\textwidth}{!}{
    \begin{tabular}{c|ccc}
        \toprule
        Method   & 1/16 (186) & 1/8 (372) & 1/4 (744) \\
        \midrule
        SupOnly  & 67.9      & 73.5      & 75.4       \\
        \midrule
        CPS~\cite{cps}     & 74.7     & 77.6     & 79.2      \\
        AEL~\cite{ael}      & 75.8     & \underline{77.9}     & 79.0      \\
        PS-MT~\cite{ps-mt}    & -        & 76.9     & 77.6       \\
        U$^{2}$\text{PL}~\cite{u2pl}     & 74.9     & 76.5     & 78.5      \\
        UniMatch~\cite{unimatch} & 75.7      & 77.3      & 78.7       \\
        \midrule
        Ours(V3)     & \underline{76.3}      & 77.8      & \underline{79.4}       \\
        Ours(B4)    & \textbf{77.4}      & \textbf{79.2}       & \textbf{80.4}   \\
        \bottomrule
    \end{tabular}
    }
    \label{tb:ex2}
\end{table}

\begin{table}[!t]
\centering
\caption{Ablation study for adaptive class-wise over-sampling (AOS). Baseline is UniMatch performance. ROS is random over-sampling. SOS is statistically based over-sampling.}
\vspace{0.1cm}
\resizebox{0.45\textwidth}{!}{
    \begin{tabular}{c|c|cc}
        \toprule
        Network                    & Sampling        & Full (1464) & 1/2(732) \\
        \midrule
                                    & Baseline        & 80.8       & 79.9     \\
                                    & ROS          & 81.2       & 80.0     \\
        DeepLabV3+~\cite{deeplabv3+}                  & SOS           & 80.9       & 80.1      \\
                                    & Ours(AOS)             & \underline{81.7}       & \underline{80.6}     \\
        \midrule
        SegFormer~\cite{segformer}                & Ours(AOS)              & \textbf{82.8}        & \textbf{81.3}    \\
        \bottomrule
    \end{tabular}
    }
    \label{tb:ex3}
\end{table}
\paragraph{Cityscapes Results}
~\tableref{tb:ex2} compares our method to the existing method on the Cityscapes dataset.
CAFS outperformed all previous methods, even on the cityscapes dataset. 
Although the previous methods produced different state-of-the-art results depending on the data partition, we achieved state-of-the-art methods by a large margin in all data partitions.

\begin{table*}[t]
\centering
\caption{Comparison with the state-of-the-art for the extended PASCAL VOC 2012 dataset. † : This means that we reproduced the approach.}
\vspace{0.1cm}
\resizebox{0.8\textwidth}{!}{
\normalsize
\begin{threeparttable}
    \def\arraystretch{1.0}
        \begin{tabular}{c|c|cc|cccc}
        \toprule
        Method   & Resolution & Network    & Backbone           & 1/16 (662)        & 1/8 (1323)         & 1/4 (2646)         & 1/2 (5291)         \\
        \midrule
        SupOnly  & 321 $\times$ 321    & DLV3+~\cite{deeplabv3+} & R50 / R101~\cite{resnet} & 61.2 / 65.6   & 67.3 / 70.4   & 70.8 / 72.8   & 74.9 / 76.0 \\
        CAC~\cite{cac}      & 320 $\times$ 320    & DLV3+~\cite{deeplabv3+} & R50 / R101~\cite{resnet} & 70.1 / 72.4   & 72.4 / 74.6   & 74.0 / 76.3   & -           \\
        ST++~\cite{st++}     & 321 $\times$ 321    & DLV3+~\cite{deeplabv3+} & R50 / R101~\cite{resnet} & 72.6 / 74.5   & 74.4 / 76.3   & 75.4 / 76.6   & -           \\
        UniMatch$\dagger$~\cite{unimatch} & 321 $\times$ 321    & DLV3+~\cite{deeplabv3+} & R50 / R101~\cite{resnet} & 74.5 / 76.5   & 75.8 / 77.0   & 76.1 / 77.2   & 77.1 / 77.5 \\
        \midrule
        Ours(V3) & 321 $\times$ 321    & DLV3+~\cite{deeplabv3+} & R50 / R101~\cite{resnet} & 75.6 / \underline{77.1} & 76.5 / \underline{77.5} & 76.8 / \underline{78.0} & 77.7 / \underline{78.1} \\
        Ours(B4) & 321 $\times$ 321    & SegF.~\cite{segformer}  & MiT-B4~\cite{segformer}             & \textbf{77.5}       & \textbf{78.8}       & \textbf{79.9}       & \textbf{80.2}       \\
        \bottomrule
        \toprule
        SupOnly  & 513 $\times$ 513    & DLV3+~\cite{deeplabv3+} & R50 / R101~\cite{resnet} & 62.4 / 67.5   & 68.2 / 71.1   & 72.3 / 74.2   & - \\
        CPS~\cite{cps}      & 512 $\times$ 512    & DLV3+~\cite{deeplabv3+} & R50 / R101~\cite{resnet} & 72.0 / 74.5 & 73.7 / 76.4 & 74.9 / 77.7 & 76.2 / 78.6 \\
        ELN~\cite{eln}      & 512 $\times$ 512    & DLV3+~\cite{deeplabv3+} & R50 / R101~\cite{resnet} & -           & 73.2 / 75.1  & 74.6 / 76.6  & -           \\
        U$^{2}$\text{PL}$\dagger$~\cite{u2pl}     & 513 $\times$ 513    & DLV3+~\cite{deeplabv3+} & R50 / R101~\cite{resnet} & 72.0 / 74.4   & 75.1 /  77.6   & 76.2 /  78.7   & \,\;\;- \,\;\;/ \underline{80.5}      \\
        PS-MT~\cite{ps-mt}    & 512 $\times$ 512    & DLV3+~\cite{deeplabv3+} & R50 / R101~\cite{resnet} & 72.8 / 75.5   & 75.7 / 78.2   & 76.4 / 78.7   & 77.9 / 79.8 \\
        UniMatch$\dagger$~\cite{unimatch} & 513 $\times$ 513    & DLV3+~\cite{deeplabv3+} & R50 / R101~\cite{resnet} & 75.8 / 78.1   & 76.9 / 78.4   & 76.8 / 79.2   & 77.7 / 79.4 \\
        \midrule
        Ours(V3) & 513 $\times$ 513    & DLV3+~\cite{deeplabv3+} & R50 / R101~\cite{resnet} & 76.1 / \underline{78.3}        & 77.3 / \underline{78.5}        & 77.4 / \underline{79.2}        & 78.4 / 80.2        \\
        Ours(B4) & 513 $\times$ 513    & SegF.~\cite{segformer}  & MiT-B4~\cite{segformer}             & \textbf{78.4}        & \textbf{78.9}        & \textbf{80.0}        & \textbf{80.7}       \\
        \bottomrule
        \end{tabular}
\end{threeparttable}
}
\label{tb:ex4}
\end{table*}

\subsection{Ablation Studies}
In this section, we examine the effects of the proposed AOS and ACT on the PASCAL VOC 2012 dataset.
In addition, we evaluated different backbones and resolutions to confirm that our CAFS was effective in extending PASCAL VOC 2012.

\paragraph{Effectiveness of Adaptive class-wise Oversampling}
~\tableref{tb:ex3} shows the results of comparing AOS and the dataset random oversampling (ROS) and statistics-based over-sampling (SOS) (Effect of increasing simple training itera-tions).
ROS is a method of oversampling that selects a random image with random oversampling.
SOS is statistical based oversampling, which is a method of oversampling by selecting an image that contains a class that is present in a small amount throughout the dataset.
As shown in ~\tableref{tb:ex3}, our AOS achieves a higher performance than the widely used statistical oversampling method.
In addition, statistical-based oversampling did not significantly improve the performance compared to the baseline method.
These experimental results indicate that oversampling relies on the difficulty of the class rather than the number of classes.

\paragraph{Effectiveness of Adaptive class-wise Confidence Threshold}
~\tableref{tab:ex5} is an ablation study of the effective confidence threshold value as a function of data partition.
We conducted experiments with (full) and (1/8) data partitions to determine the best ACT ranges.
We obtained the best performance settings in DeepLabV3+ by setting the minimum and maximum confidence thresholds to 0.85 and 0.95 in the full dataset experiment.
Comparison of two columns with the full and 1/8 datasets shows that the performance of the model depends significantly on the number of labeled samples for training.
In the 1/8 labeled partition, setting the minimum and maximum thresholds to 0.95 and 0.98 achieves the best mIoU score.
These results demonstrate the need for an adaptively set confidence threshold.

\begin{table}[h]
\centering
\caption{Ablation study of adaptive class-wise confidence threshold (ACT). This suggests that a range of different confidence thresholds should be used depending on the number of labeled data.}
\vspace{0.1cm}
\resizebox{0.45\textwidth}{!}{
{\scriptsize
    \normalsize
    \def\arraystretch{0.9}
        \begin{tabular}{c|c|cc}
        \toprule
        Network                    & $min_{ct}$$\sim$$max_{ct}$ & Full (1464) & 1/8 (183) \\
        \midrule
                                    & Baseline     & 80.8       & 77.1     \\
                                    & 0.85$\sim$0.95   & \underline{81.3}       & 75.1      \\
        DeepLabV3+~\cite{deeplabv3+}                  & 0.85$\sim$0.98   & 80.5       & 76.9     \\
                                    & 0.90$\sim$0.95   & 81.1       & 74.3      \\
                                    & 0.95$\sim$0.98   & 80.5       & \underline{77.5}     \\
        \midrule
        SegFormer~\cite{segformer}                & best range         & \textbf{82.5}       & \textbf{77.8}     \\
        \bottomrule
        \end{tabular}
    }
    }
    \label{tab:ex5}
\end{table}

\begin{table}[h]
\centering
\caption{Ablation study on the effectiveness of various components in CAFS.}
\vspace{0.1cm}
\resizebox{0.48\textwidth}{!}{
    \def\arraystretch{0.9}
        \begin{tabular}{ccc|ccccc}
        \toprule
        AOS         & ACT & Network & full(1464) & 1/2(732) & 1/4(366)  & 1/8(183)  & 1/16(92)\\
        \midrule
            -       &     -         &     -        & 80.8  & 79.9 & 78.5 & 77.1 & 74.2    \\
        \checkmark &    -          &    -         & 81.7  & 80.6 & 78.9  & 77.5 & 75.1 \\
           -        & \checkmark   &    -         & 81.3  & 80.4 & 78.9  & 77.5 & 75.2 \\
        \checkmark & \checkmark   &     -        & 82.1   & 80.7 & 79.2  & 77.7 & 75.2    \\ \hline
             -      &       -       & \checkmark  & 82.2  & 80.8 & 79.6  & 77.5 & 73.9  \\
        \checkmark &       -       & \checkmark  & 82.8   & 81.3 & 80.0  & 77.9 & 74.3 \\
             -      & \checkmark   & \checkmark  & 82.5   & 81.2 & 80.0  & 77.8 & 74.3  \\
        \checkmark & \checkmark   & \checkmark  & \textbf{83.0}  & \textbf{81.5} & \textbf{80.3}  & \textbf{78.5}  & \textbf{75.2}  \\
        \bottomrule
        \end{tabular}
        }
\label{tab:ex6}
\end{table}

\paragraph{Effectiveness of Components}
~\tableref{tab:ex6} shows an ablation study of the effects of AOS and ACT on the mIoU score in DeepLabV3+ and SegFormer. 
As shown in the table, AOS and ACT achieved performance improvements for both DeepLabV3+ and SegFormer. 
A single AOS improved mIoU from 80.8\% to 81.7\% and 78.5\% to 78.9\% in full and 1/4 partitions, respectively.
A single ACT achieved improvements of 0.5\% and 0.3\%, respectively.
The combination of AOS and ACT significantly outperformed the baseline by 1.3\% and 0.7\% mIoU gaps compared to the baseline.
Switching the backbone network from DeepLabV3+ to SegFormer leads to performance improvement in all partitions.

\begin{figure*}[!t]
\centering
  \includegraphics[width=0.95\linewidth]{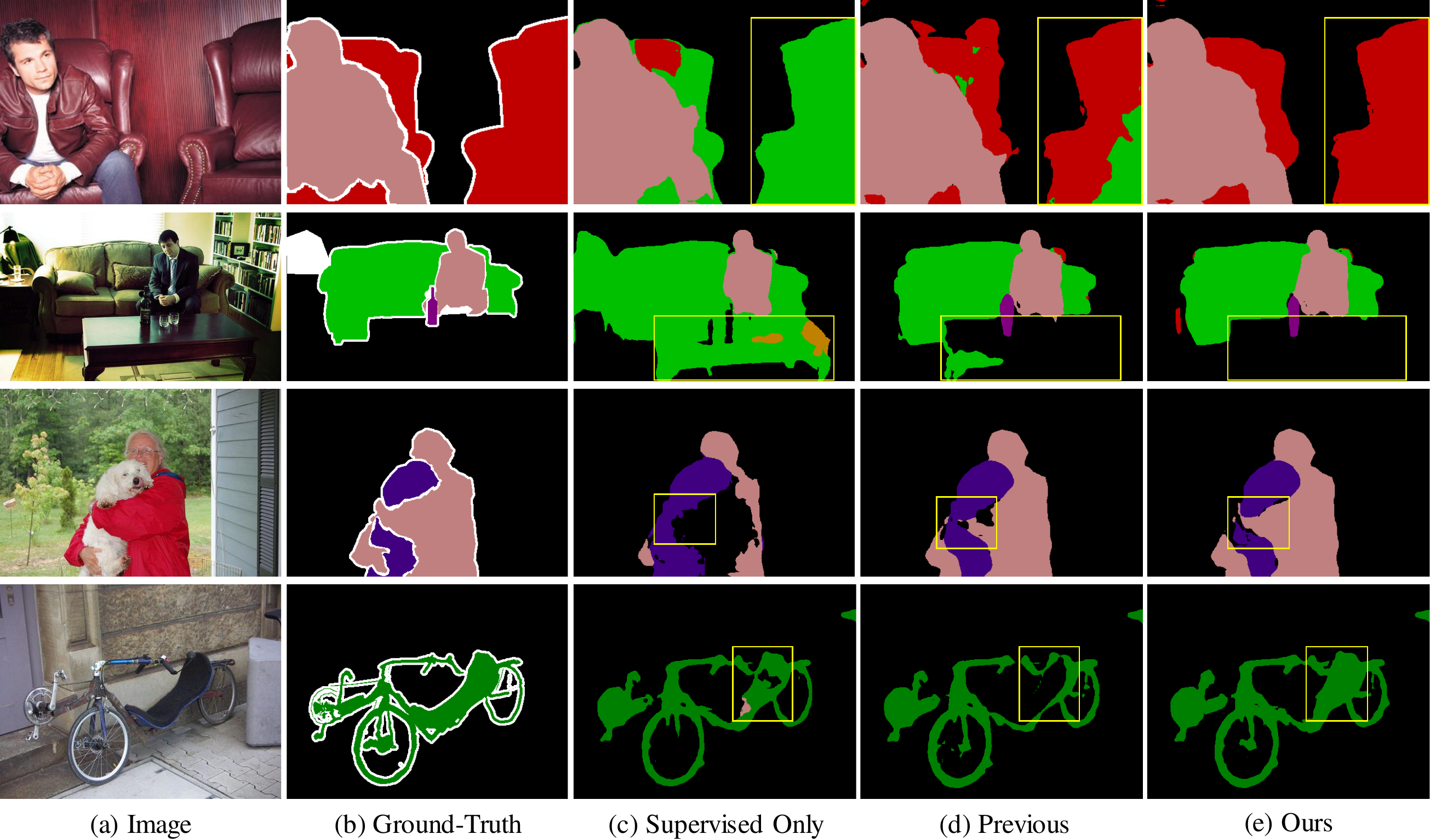}
  \caption{Qualitative results for the PASCAL VOC 2012 val set. All results are models trained on the full data partition of the PASCAL VOC 2012 dataset using the DeepLabV3+ network and ResNet101 backbone. The images in each column are (a) input image, (b) ground truth image, (c) predictions from models trained only under supervision on labeled data without unlabeled data, (d) predictions from UniMatch, an earlier state-of-the-art method, (e) predictions from our CAFS. The yellow rectangles show that CAFS achieves better segmentation results by training more pseudo-labels with high quality and classes with poor performance.}
\label{fig:pq2}
\end{figure*}

\paragraph{Extended PASCAL VOC 2012 Results}
~\tableref{tb:ex4} shows the results of evaluating whether CAFS is valid not only in finely annotated datasets such as Cityscapes and PASCAL VOC 2012, but also in coarsely annotated datasets.

\begin{figure}[!t]
\centering
  \includegraphics[width=\linewidth]{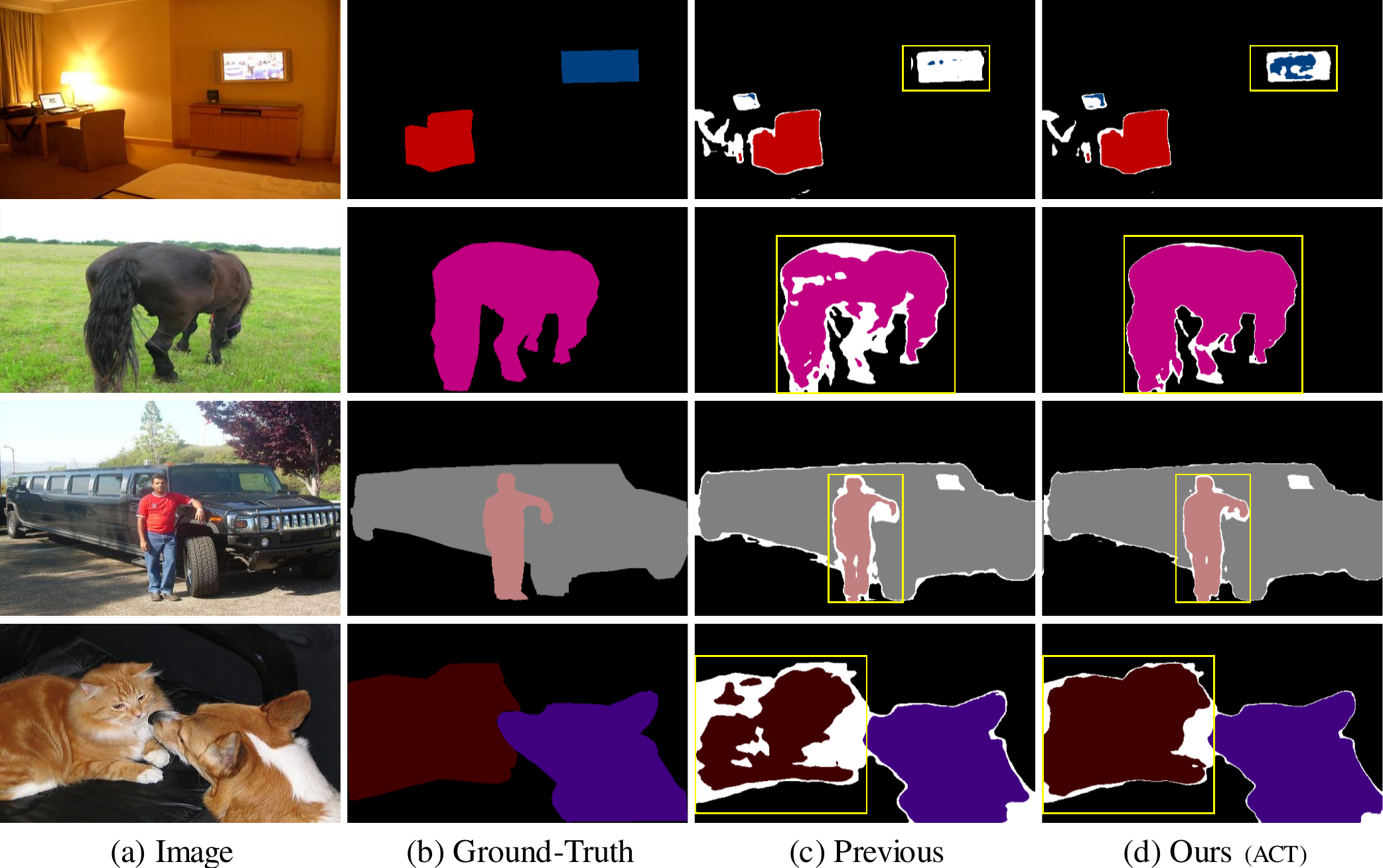}
  \caption{Pseudo-label qualitative results. Use the images from the PASCAL VOC 2012 unlabeled set. All results are models trained on the full data partition of the PASCAL VOC 2012 dataset using the DeepLabV3+ network and ResNet101 backbone. (a) input image, (b) ground truth image, (c) pseudo-label from UniMatch with a confidence threshold of 0.95 for all classes, (d) pseudo-label using ACT. The yellow rectangles highlight that the ACT contains more information when creating the pseudo-label.}
\label{fig:pq}
\vspace{-0.2cm} 
\end{figure}

We evaluated CAFS with different backbones and training resolutions to assess scalability.
Our CAFS outperformed all previous methods and achieved state-of-the-art performance in all backbones and models. 
These experimental results show that the proposed CAFS has high scalability for semi-supervised semantic segmentation.

\subsection{Qualitative Results}
In ~\figref{fig:pq}, we qualitatively compare the quality of pseudo label between the previous state-of-the-art approach, Uni-Match, and the proposed CAFS. 
As can be seen in the figure, more true pixels are intact in our CAFS, and CAFS generates pseudo-label classes adaptively under the influence of ACT. 
Compared to CAFS, the previous state-of-the-art method, UniMatch, causes numerous true pixel information losses and is confused by similar classes because it assumes fixed confidence for all classes.
Therefore, the results of this qualitative analysis suggests that CAFS better captures fine-grained contextual information and CAFS explicitly discriminates the confounding classes with the two core class-adaptive operations, AOS and ACT.

~\figref{fig:pq2} shows the qualitative results for hard classes in the PASCAL VOC 2012 dataset.
As shown in the figure, CAFS shows performance improvement in hard classes under the influence of AOS and ACT based on validation performance analysis.
These experimental results show that CAFS alleviates the performance imbalance problem.

\label{sec:exp}

\section{Conclusions}
In this paper, we proposed CAFS that solves the problem of high and fixed confidence-based pseudo-labeling methods. 
We experimentally show that the two core operations, AOS and ACT, adaptively adjust the trustworthy confidence threshold using the calibrated scores.
Our ablation study showed their efficiency with different numbers of labeled samples, backbones, and the effectiveness of ACT and AOS.

\label{sec:con}

\appendix
\section{Appendix}
\subsection{Training Curves}
\begin{figure}[ht!]
  \centering
  \includegraphics[width=1.0\linewidth]{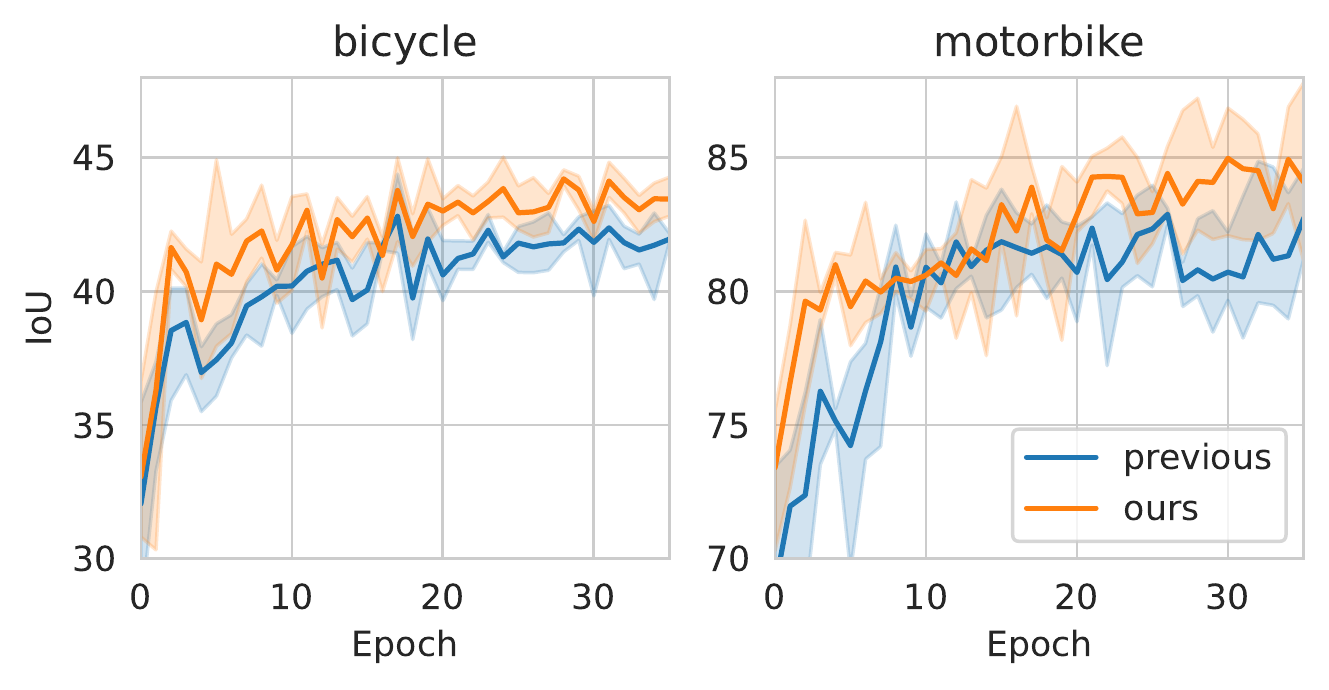}
  \caption{ Comparison of training curves between our CAFS and the previous state-of-the-art, UniMatch. The extended PASCAL VOC 2012 in 1/4 (2,646) data partition, we compare \textit{bicycle} and \textit{motorbike} classes, which are confused with each other.}
  \label{fig:sup_fig1}
  \vspace*{-0.3cm}
\end{figure}

~\figureref{fig:sup_fig1} is a training curve when CAFS can only be applied on a bicycle.
Our intended bicycle class can see an improvement in performance compared to when it was not applied.
In addition, it was found that the performance of motorcycles, which are often confused with the bicycle class, was improved.
These experimental results indicate that there is a strong interaction between classes that are often confused with each other, and it is expected that additional performance improvement can be achieved by considering these factors when determining oversampling and threshold.

\begin{figure}[ht!]
  \centering
  \includegraphics[width=1.0\linewidth]{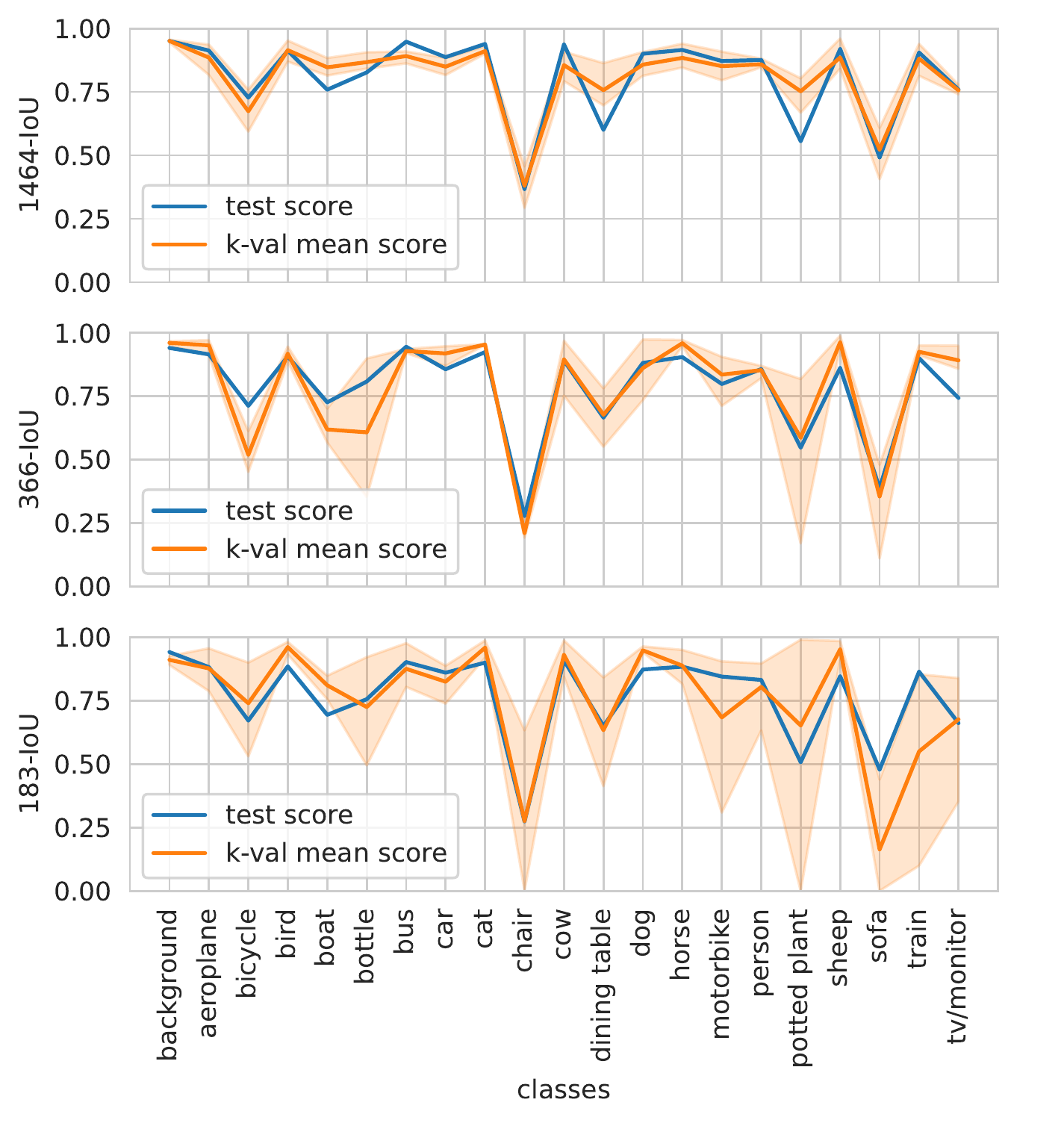}
  \caption{ Comparison of performance curves between K-validation average class-wise IoU and test class-wise IoU. The orange line shows the average class-wise IoU of k validation sets and the blue line shows the class-wise IoU of the test set. For comparison, we used full (1464), 1/4 (366), and 1/8 (183) data partitions of  base PASCAL VOC 2012.}
  \label{fig:sup_fig5}
  \vspace*{-0.3cm}
\end{figure}

\subsection{Class-wise Performance Curves}
Fig.~\ref{fig:sup_fig5} shows a graph comparing the K-validation average class-wise IoU to the test class-wise IoU. 
Comparing the class-wise IoU score with single set of validation, i.e., K=1, to the test class-wise IoU, we can see that the performance deviation is larger with less data.
This performance deviation suggests that using only single validation is not representative of the data.
Therefore, we configure K-validation to reduce the performance deviation.
The results in Fig.~\ref{fig:sup_fig5} show that the k validations average class-wise IoU and test set IoU have similar performance trends.
These experimental results indicate that a diverse validation set partitioning can better represent the data than a single validation set, and we expect that it can further stabilize our proposed AOS and ACT.

\subsection{Additional Qualitative Results}
\begin{figure*}[t!]
  \centering
  \includegraphics[width=\linewidth]{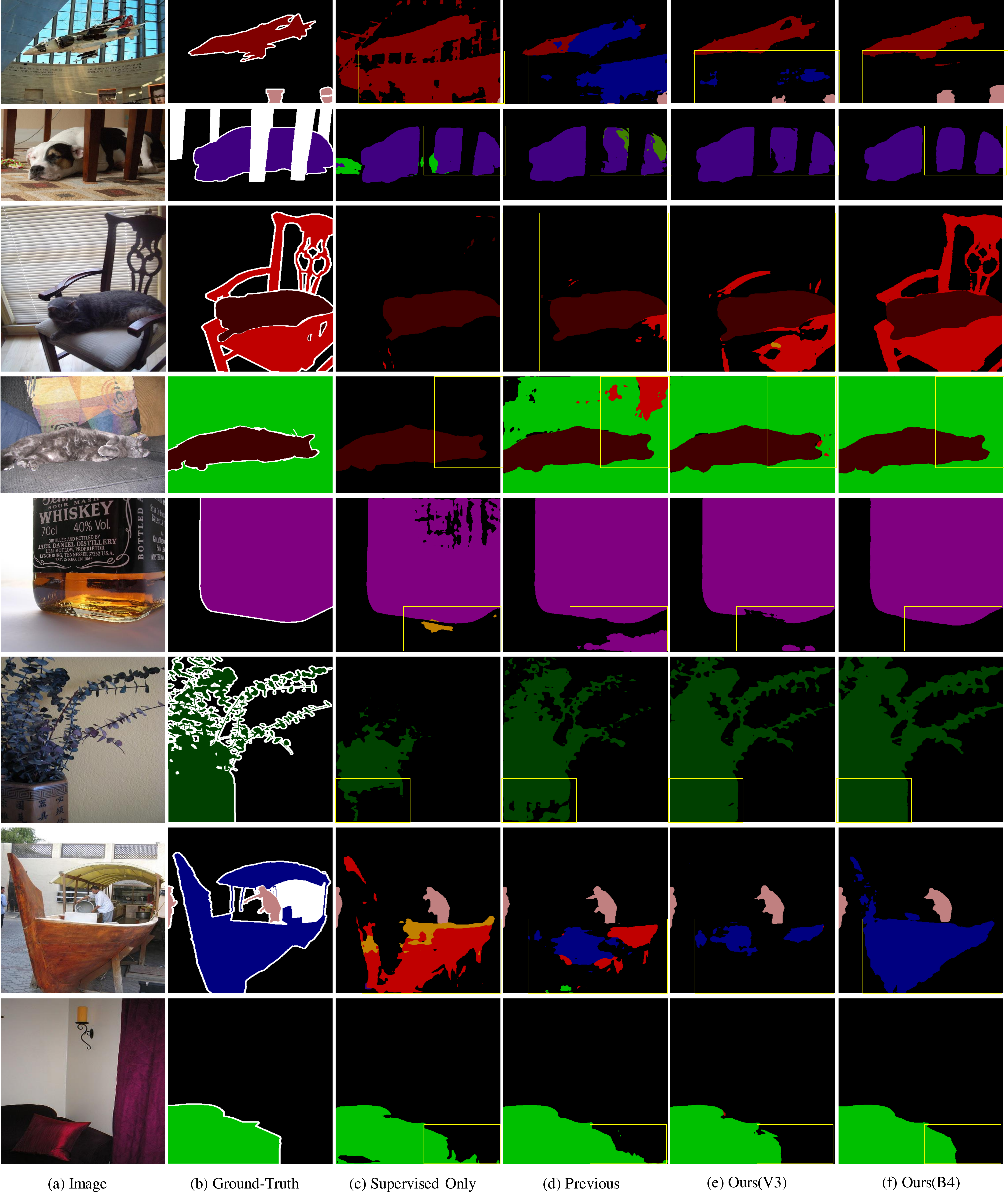}
  \caption{Qualitative results for the \textbf{PASCAL VOC 2012} val set. The images in each column are (a) input image, (b) ground truth image, (c) predictions of models trained only supervised on labeled data without unlabeled data, (d) predictions from UniMatch, a previously state-of-the-art method, (e) predictions from our CAFS using DeepLabV3+ network, (f) predictions from our CAFS using Segformer network. Yellow rectangles highlight that CAFS produces better segmentation results.}
  \label{fig:sup_fig3}
  \vspace*{-0.2cm}
\end{figure*}

\begin{figure*}[t]
  \centering
  \includegraphics[width=\linewidth]{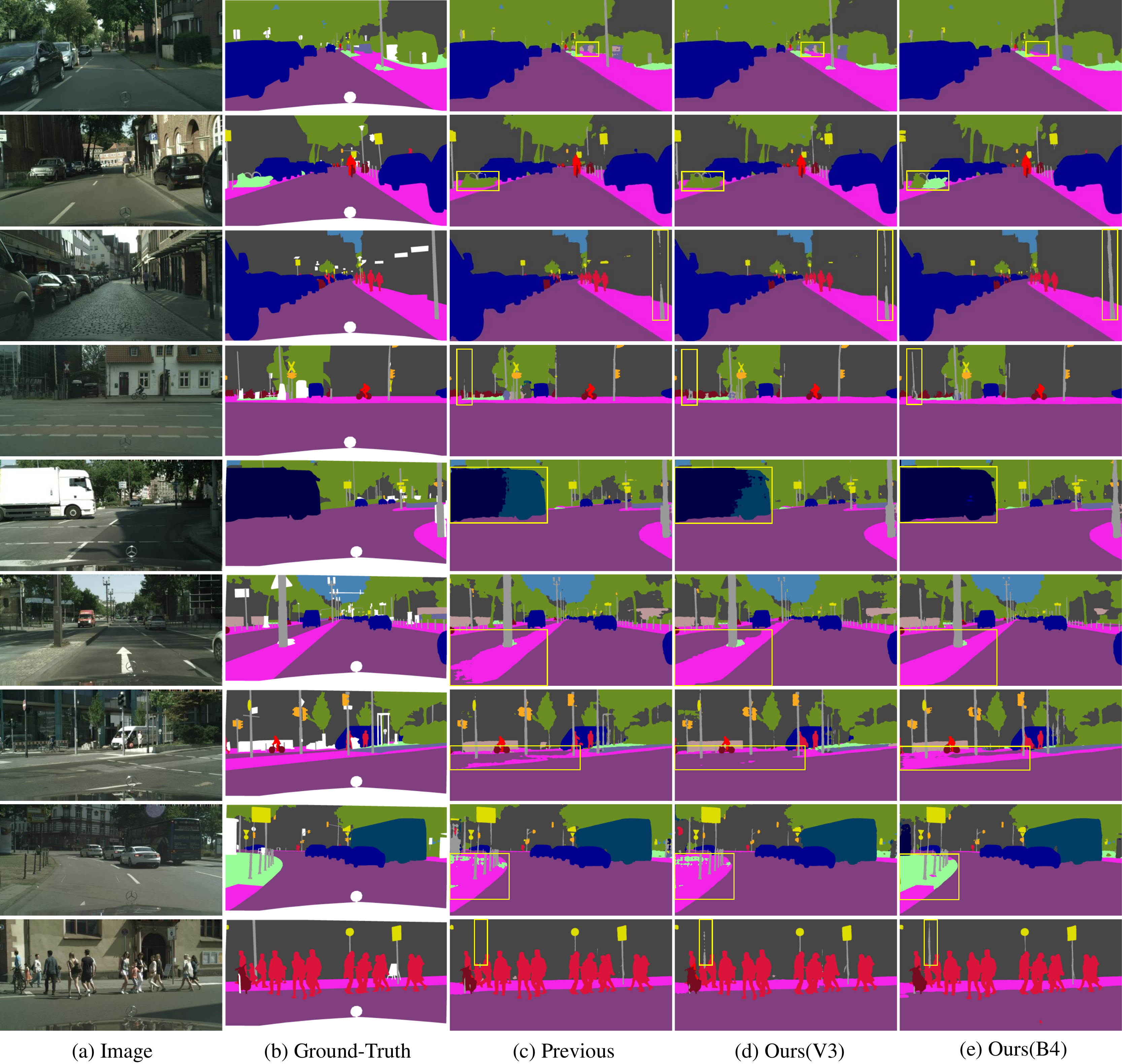}
  \caption{Qualitative results for the \textbf{Cityscapes} val set. The images in each column are (a) input image, (b) ground truth image, (c) predictions from UniMatch, a previously state-of-the-art method, (d) predictions from our CAFS using DeepLabV3+ network, (e) predictions from our CAFS using Segformer network. Yellow rectangles highlight that CAFS produces better segmentation results.}
  \label{fig:sup_fig4}
  \vspace*{-0.2cm}
\end{figure*}

Fig.~\ref{fig:sup_fig3} and Fig.~\ref{fig:sup_fig4} show a qualitative comparison on the base PASCAL VOC 2012~\cite{pascal} and Cityscapes~\cite{cityscapes} datasets, respectively.
The yellow boxes highlight the areas where we improved performance.
Our proposed CAFS shows better segmentation results than the existing state-of-the-art method.

\subsection{Training Details of CAFS}
In PASCAL VOC 2012, we use the SGD optimizer with momentum of 0.9 and weight decay of $1 \times 10^{-4}$. We use learning rates of 0.001 and 0.01 for the encoder and decoder, and we train using 80 epochs with a poly learning rate scheduler. The lambda of AOS hyperparameter uses 10 and the class-wise precision threshold $max_{cp}$ is set to 0.95. In this dataset, both DeepLabV3+~\cite{deeplabv3+} and SegFormer~\cite{segformer} use the same settings.

Cityscapes use different settings for DeepLabV3+ and Segformer. In DeeplabV3+, the learning rate of the encoder is 5e-3 and the learning rate of the decoder is 5e-2, and we use a poly learning rate scheduler to train 240 epochs. Segformer uses the AdamW~\cite{adamw} optimizer with weight decay of 1e-2. The encoder has a learning rate of 6e-5, the decoder has a learning rate of 6e-4, and uses a poly learning rate scheduler to train 240 epochs. Both networks use 8 GPUs and other settings such as AOS hyperparameter lambda and class-wise precision threshold are identical to the PASCAL VOC settings.

%

%%%%%%%%% REFERENCES
{\small
\bibliographystyle{ieee_fullname}
\bibliography{egbib}
}

\end{document}